\documentclass[twoside,11pt]{article}

\usepackage{jmlr2e}
\usepackage{graphicx}
\usepackage{caption}
\usepackage{amsmath}
\usepackage{algorithm}
\usepackage[noend]{algpseudocode}
\usepackage{array}
\usepackage{pbox}

\usepackage{hyperref}

\ShortHeadings{Application of Self-Play Reinforcement Learning to a Four-Player Game of Imperfect Information}{Charlesworth}
\firstpageno{1}

\begin{document}

\title{Application of Self-Play Reinforcement Learning to a Four-Player Game of Imperfect Information}

\author{\name Henry Charlesworth \email H.Charlesworth@warwick.ac.uk \\
       \addr Centre for Complexity Science\\
       University of Warwick}

\editor{}

\maketitle

\begin{abstract}
We introduce a new virtual environment for simulating a card game known as ``Big 2". This is a four-player game of imperfect information with a relatively complicated action space (being allowed to play 1,2,3,4 or 5 card combinations from an initial starting hand of 13 cards). As such it poses a challenge for many current reinforcement learning methods. We then use the recently proposed ``Proximal Policy Optimization" algorithm \citep{ppo} to train a deep neural network to play the game, purely learning via self-play, and find that it is able to reach a level which outperforms amateur human players after only a relatively short amount of training time.
\end{abstract}


\section{Introduction}
Big 2 (also known as deuces, big deuce and various other names) is a four player card game of Chinese origin which is played widely throughout East and South East Asia. The game begins with a standard deck of 52 playing cards being shuffled and dealt out so that each player starts with 13 cards. Players then take it in turns to either play a hand or pass with basic aim of being the first player to be able to discard all of their cards (see section 2 for more details about the rules). In this work we introduce a virtual environment to simulate the game which is ideal for the application of multi-agent reinforcement learning algorithms. We then go on to train a deep neural network which learns how to play the game using only self-play reinforcement learning. This is an interesting environment to study because the most remarkable successes that have come from self-play reinforcement learning such as Alpha Go \citep{alphago} and Alpha Zero \citep{alphazero} have been confined to two-player games of perfect information (e.g. Go, Chess and Shogi). In contrast Big 2 is a four-player game of imperfect information where each player is not aware of the cards that are held by the other players and so does not have access to a full description of the game's current state. In addition to this Alpha Zero supplements its training and final decision making with a monte carlo tree search which requires the simulation of a large number of future game states in order to make a single decision whereas here we consider only training a neural network to make its decision using the current game state that it receives. This is also in contrast to the most successful Poker playing programs such as Libratus \citep{libratus} and DeepStack \citep{deepstack} which again require much more computationally intense calculations to perform at the level that they do (e.g. DeepStack uses a heuristic search method adapted to imperfect information games). One approach which does directly apply deep self-play reinforcement learning to games of imperfect information is ``neural fictitious self-play" \citep{nfsp} where an attempt is made to learn a strategy which approximates a Nash equilibrium, although this has not been applied to any games with more than two players.

Multi-agent environments in general pose an interesting challenge for reinforcement learning algorithms and many of the techniques which work well for single-agent environments cannot be readily adapted to the multi-agent domain \citep{multiagentactorcritic}. Approaches such as Deep Q-Networks \citep{deepqnetworks} struggle because multi-agent environments are inherently non-stationary (due to the fact that the other agents are themselves improving with time) which prevents the straightforward use of experience replay that is necessary to stabilize the algorithm. Standard policy gradient methods also struggle due to the large variances in gradient estimates that arise in the multi-agent setting which often increase exponentially with the number of agents. Although there are some environments that are useful for testing out multi-agent reinforcement learning algorithms such as the OpenAI competitive environments \citep{multiagentcompetition} and the ``Pommerman" competitions (\url{https://pommerman.com}) we hope that Big 2 can be a useful addition as it is relatively accessible whilst still requiring complex strategies and reasoning to play well.

\section{Rules and Basic Strategy}
At the start of each game a standard deck of playing cards (excluding jokers) is dealt out randomly such that each of the four players starts with 13 cards. The "value" of each card is ordered primarily by number with 3 being the lowest and 2 being the highest (hence Big 2), i.e. $ 3<4<5<6<7<8<9<10<\text{J}<\text{Q}<\text{K}<\text{A}<2 $ and then secondly by suit with the following order: $ \text{Diamonds} < \text{Clubs} < \text{Hearts} < \text{Spades}$. Throughout the rest of the paper we will refer to cards by their number and the first letter of their suit, so for example the four of hearts will be referred to as the 4H. This means that the 3D is the lowest card in the game whilst the 2S is the highest. There are a number of variations in the rules that are played around the world but in the version which we use the player who starts with the 3D has to play this card first as a single. The next player (clockwise) then either has to play a higher single card or pass, and this continues until either each player passes or someone plays the 2S. At this point the last player to have played a card is ``in control" and can choose to play any single card or \textit{any valid poker hand}. These include pairs (two cards of the same number), three-of-a-kinds (three cards of the same number), four-of-a-kinds (four cards of the same number), two-pairs, straights (5 cards in numerical order, e.g. $8,9,10,\text{J},\text{Q}$), flushes (5 cards of the same suit), full-houses (3 cards of one number, 2 of another number) and straight-flushes (both a straight and a flush). Subsequent players must then either play a better hand \textit{of the same number of cards} or pass. This continues until everyone passes at which point the last player gets control and can again choose to play any valid hand they wish. The game finishes once one player has gotten rid of all of their cards at which point they are awarded a positive reward equal to the sum of the number of cards that the three other players have left. Each of the other players is given a negative reward equal to the number of cards they have left - so for example if player 1 wins and players 2,3 and 4 have 5, 7 and 10 cards left respectively then the rewards assigned will be $ \{ 22, -5, -7, -10 \}$. This provides reasonable motivation to play to win in most situations rather than just trying to get down to having a low number of cards left.

\begin{figure}[!htb]
\centering
\fbox{\includegraphics[width=0.8\textwidth]{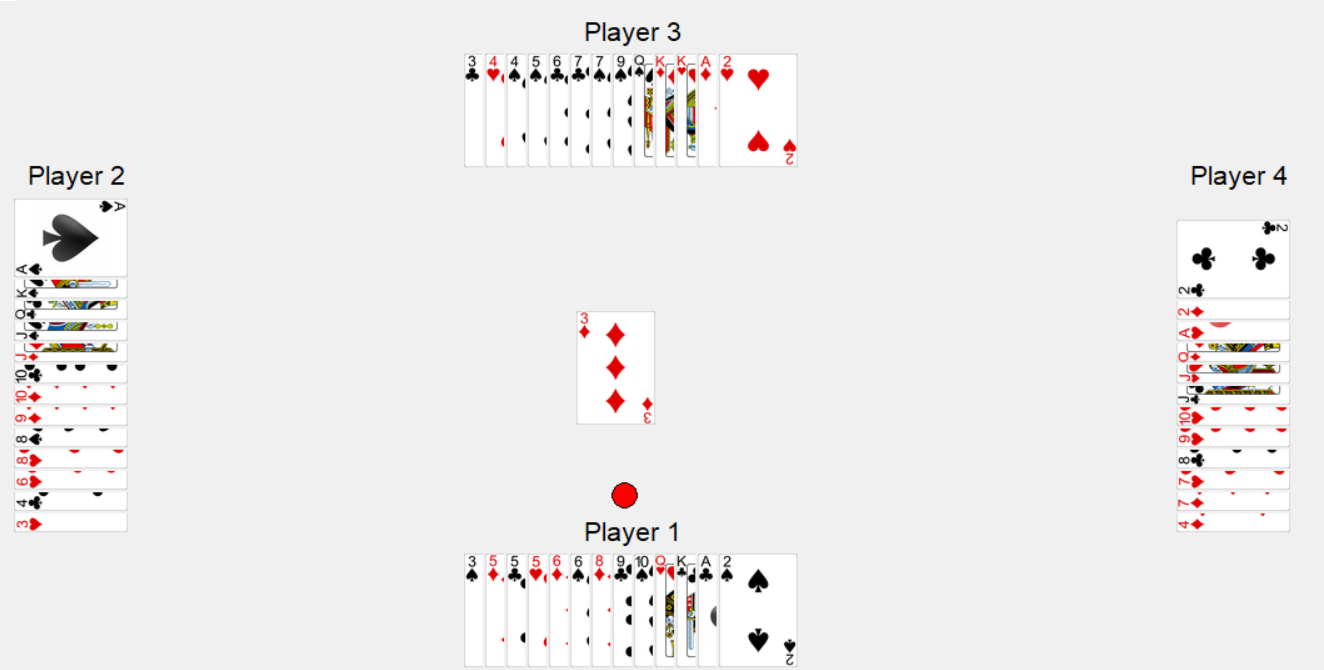}}
\caption{A typical start to a game (although note that players are not aware of what cards the other players hold). All 52 cards are dealt out so that each player begins with 13 cards. The player with the 3 of diamonds (here player 4) must start and play this as a single card hand. Subsequent players must play a higher single card or pass (skip their go). This continues until everyone passes at which point the last player who played a card gains ``control". A player with control can then choose to play any valid 1,2,3,4 or 5 card hand (see text for details). Subsequent players must then play a better hand of the \textit{same number of cards} or pass until someone new gains control. This continues until one player has managed to play all of their cards.}
\end{figure}

In terms of hand comparisons for hands that consist of more than one card we have the following rules: two-card hands (i.e. pairs) are ranked primarily on number such that e.g. $ [5x, 5y] < [10w, 10z]$ regardless of suits and then secondly on suit (the pair containing the highest suit is higher, e.g. $ [10\text{C}, 10\text{H}] < [10\text{D}, 10\text{S}]$). For three card hands only the number is important (as you never have to compare two three card hands of the same number). For four card hands when we compare two-pairs only the highest pair is important (so e.g. $[\text{QD}, \text{QS}, \text{JH}, \text{JS}] < [\text{KC}, \text{KH}, 4\text{C}, 4\text{H}]$) and a four-of-a-kind beats any two-pair. For five card hands we have that: $\text{Straight} < \text{Flush} < \text{Full House} < \text{Straight Flush}$. If we are comparing straights then whichever one contains the largest individual single card will win and the same goes for comparing two flushes. Full houses are compared based on the number which appears three times in it, so for example: $ [2\text{S}, 2\text{H}, 5\text{C}, 5\text{H}, 5\text{S}] < [3\text{S}, 3\text{H}, 10\text{H}, 10\text{S}, 10\text{C}]$.

The skill of the game is in coming up with a plausible strategy for being able to play all of one's cards. This often needs to be adapted as a result of the strategies which one's opponents play and includes identifying situations when the chances of winning are so low that it is best to try and aim for ending with a low number of cards rather than actually playing to win. This involves knowing when to save hands for later that one could play immediately but which might turn out to be a lot more useful at a later stage of the game. Whilst there is certainly a significant amount of luck involved in terms of the initial hand that one is dealt (such that the result of any individual game shouldn't be taken to be too meaningful) if one plays against more experienced opponents it will quickly become apparent that there is also a large skill component involved such that a good player will have a significant edge over a less experienced player in the long run.

\section{Virtual Big 2 Environment}
A virtual environment written in Python which simulates the game is available alongside the source code used for training the neural network to play here: \url{https://github.com/henrycharlesworth/big2_PPOalgorithm}. The environment operates in a way which is fairly similar to those which are included in OpenAI Gym \citep{openaigym} but with a few differences. The primary functions used are:
\begin{verbatim}
env = big2Game(); env.reset() #set up and reset environment.
players_go, current_state, currently_available_actions = env.getCurrentState()
reward, done, info = env.step(action) #play chosen action and update game.
\end{verbatim}

There is also a parallelized implementation of the environment included. This uses Python's multiprocessing module to run multiple different games at the same time on different cores which was particularly useful for the method we used to train a neural network to play (see next section).

\subsection{Describing the State of the Game}
One of the most important steps for being able to train a neural network to play is to determine a sensible way of encoding the current state of the game into a vector of input features. Technically a full description of the current game state would involve information about the actual hand the player has but also about every other hand that each other player has played before them as well as any potentially relevant information about what you believe the other players' styles of play to be. Given that it is possible for some games to last over 100 turns storing complete information like this would lead to potentially huge input states containing a lot of information which is not particularly important when making most decisions. As such we design an input state by hand which contains a small amount of "human knowledge" about what we deem to be important for making decisions during the game. Note that this is the only stage at which any outside knowledge about the game is built into our method for training a neural network and we have tried to keep this fairly minimal. Details about this can be found in Appendix A. 

\subsection{Representing the Possible Actions}
Modelling the available actions takes a bit more thought as generally there are many ways you can make poker hands from a random set of 13 cards and we need a systematic way of indexing these. We found that the best way to do this is to ensure that we store a player's hand sorted in order of value and then define actions in terms of the indices of the cards within the hand. So for example if we are considering actions involving five cards and a player has a hand $[ 3\text{C}, 3\text{S}, 4\text{H}, 6\text{D}, 7\text{H}, 8\text{C}, 9\text{D}, 10\text{C}, \text{KS}, \text{AC}, \text{AS}, 2\text{C}, 2\text{S} ]$ then we could define the action of playing the straight $[6\text{D}, 7\text{H}, 8\text{C}, 9\text{D}, 10\text{C}]$ in terms of the ordered card indices within the hand (using 0 as the starting index): $[3,4,5,6,7]$. If we were thinking instead the flush $[3\text{C}, 8\text{C}, 10\text{C}, \text{AC}, 2\text{C}]$ this can be defined by its card indices $[0,5,7,9,11]$. This is fine because the input state to the neural network tells us about which card value actually occupies each of the card indices in the current hand. We can then construct look up tables that convert between card indices and a unique action index (see Appendix B for details and some pseudocode). Doing this we find that there are a total of 1695 different moves that could potentially be available in any given state, although a majority of time the actual number of allowed moves will be significantly lower than this.

\section{Training a Network Using Self-Play Reinforcement Learning}
To train a neural network to play the game we make use of the ``Proximal Policy Optimization" (PPO) algorithm proposed recently by \citet{ppo} which has been shown to inherit the impressive robustness and sample efficiency of Trust Region Policy Optimization methods \citep{trpo} whilst being much simpler to implement. It has also been shown to be successful in a variety of reasonably complicated competitive two-player environments such as ``Sumo" and ``Kick and Defend" \citep{multiagentcompetition} where huge batches (generated by running many of the environments in parallel) are used to overcome the problem of large variances.

The algorithm is a policy-gradient based actor-critic method in which we use a neural network to output both a policy $\pi(a|s)$ over the available actions $a$ in any given state $s$ alongside an estimate of a state value function which is used to estimate the advantage $\hat{A}(a|s)$ of taking each action in any particular state. We make use of the "generalized advantage estimation" \citep{gae} algorithm to do this. Further details of the PPO algorithm (including the hyperparameters used) and the neural network architecture can be found in Appendix C.

We then set up four copies of the current neural network (initially with random parameters) and get them to play against each other. We generate mini-batches of size 960 by running 48 separate games in parallel for 20 steps at a time. We then train for multiple epochs on each batch using stochastic gradient descent. Note that these are significantly smaller than those used in \citet{multiagentcompetition} where batches of hundreds-of-thousands were used. We then run this for $150,000,000$ total steps ($156,250$ training updates) which corresponds to approximately 3 million games. This was carried out on a single PC with four cores and a GPU and took about 2 days to complete. We did not find that it was necessary to use any kind of opponent sampling (although it would be interesting to investigate whether or not this would improve the final results) and so the neural networks were always playing the most recent copies of themselves throughout the entire duration of training. The hyperparameters we used were chosen to be similar to those which had worked previously for other tasks but interestingly we did not have to play around with any of these at all to get the algorithm to work well. It is possible we just got lucky (and we have not made any serious attempt to explore variations in hyperparameters) but this seems to back up the claim that PPO is remarkably robust.

\section{Results}
\begin{figure}
\includegraphics[width=\textwidth]{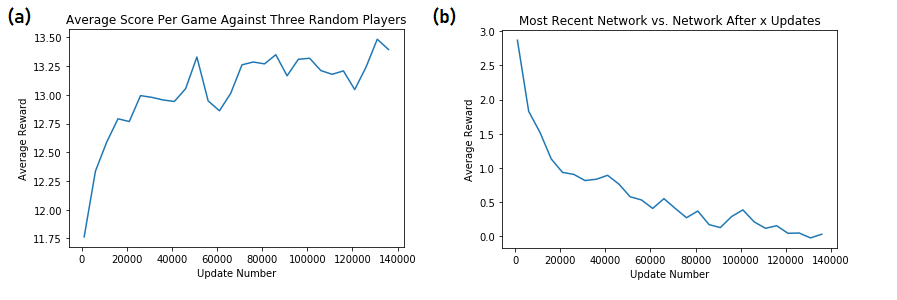}

\caption{(a) Average score per game of the trained network against three random opponents as the training progresses. (b) The final network against three copies of the network at earlier times in the training. All plotted points are averaged over 10,000 games. }
\label{fig:results}
\end{figure}

As a simple initial evaluation of the network's learning we compare how its performance against three random players progresses throughout its training (figure \ref{fig:results}(a)) as well as how it performs against earlier versions of itself (figure \ref{fig:results}(b)). Each point on these plots is averaged over $10,000$ games and the network being evaluated accounts for one player whilst the other networks (random on the left of figure \ref{fig:results} and the earlier network on the right) make up the other three players. We see it takes very little time to achieve a large positive score against random opponents and that the learning progress seems to continue steadily throughout the training (note the first point plotted is after 1000 updates, not 0). It seems likely that if left to train for longer the performance would continue to improve further.

As a more interesting test we designed a front-end to make it easy for the trained network to play against humans and recorded the results of various humans playing against three of the fully trained networks (this is available to try out for yourself at \url{https://big2-ai.herokuapp.com/game}). Although none of the players could be considered experts all of them had some experience playing the game and could be considered enthusiastic amateurs. Organizing matches against more experienced players is something we would like to arrange in the future. Full results are included in Appendix D where we see that the trained neural network significantly outperforms most of the human players.

\section{Conclusion}
In this paper we have introduced a novel environment to simulate the game of ``Big 2" in a way which is ideal for the application of multi-agent reinforcement learning algorithms. We have also been able to successfully train a neural network purely using self-play deep reinforcement learning that is able to play the game to a super-human level of performance without the need to supplement it with any kind of tree search over possible future states when making its decisions. Nevertheless it seems likely that these results can be improved upon further and so we would like to encourage anyone working on developing multi-agent learning techniques to consider trying out this environment as a benchmark.


\acks{Thanks to Liam Hawes, Katherine Broadfoot, Terri Tse, Kieran Griffiths, Shaun Fortes and James Frooms for agreeing to play competitive games against the trained network and to Professor Matthew Turner for reading this manuscript and providing valuable feedback. This work was supported by the UK Engineering and Physical Sciences Research Council (EPSRC) grant No. EP/L015374/1, CDT in Mathematics for Real-World Systems. }

\bibliography{sample}


\newpage

\appendix
\section*{Appendix A. Encoding the Current Game State}
\label{app:gamestate}
\begin{figure}
\includegraphics[width=\textwidth]{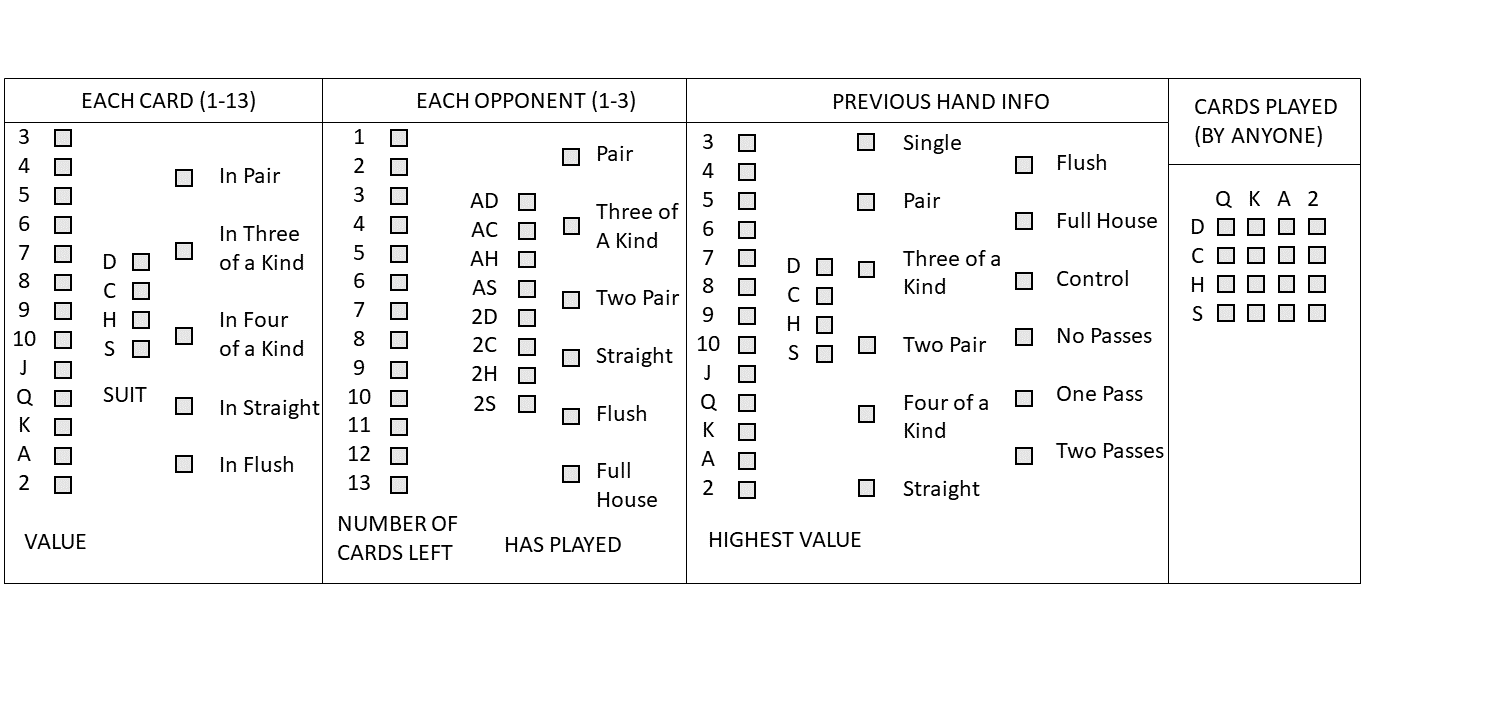}

\caption{Input state provided to the neural network which encodes the current state of the game. This includes information about the player's own hand as well as some limited information about what each of the opponents has played so far and other things which have occurred during the game up until the present point. This leads to an input of size 412 made up of zeros and ones.}
\label{fig:inputstate}
\end{figure}

Figure \ref{fig:inputstate} shows the input that is provided to the network. Firstly the player's cards are sorted into order of their value (from 3D to 2S) and labelled from 1 up to a maximum of 13. For each card in the player's current hand there are then 13 inputs that are zero or one to encode the card's value and then four more to encode the suit. As well as this we provide information about whether the card can be included in any combination of cards (i.e. is it apart of a pair, a straight etc). For each of the three opponents we keep track of the number of cards they have left as well as well as certain information about what they've played so far. In particular we keep track of whether \textit{at any point} during the game so far they've played any of the highest 8 cards (AD - 2S) as well as if they've played a pair, a two pair, a three of a kind, a straight, a flush or a full house. The network is also provided information about the previous hand which has been played (both its type and its value) as well as the number of consecutive passes made prior to the current go or if it currently has control. Finally we provide it with information about whether anyone has played any of the top 16 cards. This is potentially important for keeping track of which single is the highest left in play (and hence guaranteed to take control). We cut this at 16 to reduce the size of the input as it is rare for a high-level game to still be going when the highest cards left are lower than a queen. 

This is the way we choose to represent the current game state when training our network and also the state which is returned by the \texttt{env.step()} function in the game environment, however it also records all hands which are played in a game and so it is relatively simple to write a new function which includes more or less information if this is desired.

\section*{Appendix B. Indexing the Action Space}
\label{app:actions}
Here we give the pseudocode for generating "look-up tables" which can be used to systematically index the possible actions that are available in any given state. We consider separate look up tables for actions containing different numbers of cards. In the case of five-card hands it is possible (because of flushes) for any combination of card indices to be a valid hand meaning that under this representation there are $\binom{13}{5}=1287$ possible five-card actions. The idea is then to construct a mapping between each allowable set of indices $\{c_1, c_2, c_3, c_4, c_5\}$ and a unique action index $i$. Algorithm \ref{fivecardmoves} does this creating a matrix "actionIndices5" which can be indexed with the card indices to return $i$ and then including a reverse-look up table which maps $i$ back to the card indices. In the case of four-card actions there are constraints on the indices that can actually be used to make a valid hand because the only valid four-card hands are two pairs and four of a kinds. This means that for example the combination of indices $[2,8,9,10]$ could never be a valid hand as the cards (which are sorted in order) in positions 2 and 8 could never correspond to the same number and hence cannot be a pair. Consequently rather than there being $\binom{13}{4}=715$ possible four-card actions we find that are there are actually only 330 under this representation. Similar constraints apply to two and three card actions where we find that there are 33 and 31 possible actions respectively and then trivially there are 13 possible one-card actions. In total this gives us $1287 + 330 + 31 + 33 + 13 + 1 = 1695$ potential moves that could be allowable in any given state (the extra 1 is accounting for being allowed to pass). In the python implementation the big2Game class has a function \texttt{availAcs = big2Game.returnAvailableActions()} which returns an array of size 1695 of 0s and 1s depending on whether each potential action is actually playable for the current player in the current game state. This vector is ordered with one-card actions in indices $0-12$, two-card actions from $13-45$, three-card actions from $46-76$, four-card actions from $77-406$, five-card actions from $407-1693$ and then finally $1694$ corresponding to the pass action. The \texttt{big2Game.step(...)} function takes an action index (from $0-1694$) as its argument and \texttt{big2Game.getCurrentState()} returns as its third value a vector of 0s (corresponding to actions allowed in current state) and $-\infty$. This was just because it was convenient to use these values instead of 0s and 1s when using a softmax over the neural network output to represent the probability distribution over allowed actions but is trivial to change.

\begin{algorithm}[H]
\caption{Look up tables for five-card actions}\label{fivecardmoves}
\begin{algorithmic}[1]
\State \textbf{Initialize:} \textit{actionIndices5} as a $13 \times 13 \times 13 \times 13 \times 13$ array of zeros
\State \textbf{Initialize:} \textit{inverseIndices5} as an $1287 \times 5$ array of zeros
\State \textbf{Initialize:} $i = 0$
\For{$c_1=0$ to $8$}
	\For{$c_2=c_1+1$ to $9$}
		\For{$c_3=c_2+1$ to $10$}
			\For{$c_4=c_3+1$ to $11$}
				\For{$c_5=c_4+1$ to $12$}
					\State \textit{actionIndices5}$[c_1,c_2,c_3,c_4,c_5] = i$
					\State \textit{inverseIndices5}$[i,:] = [c_1, c_2, c_3, c_4, c_5]$
					\State $i \mathrel{+}= 1$
				\EndFor		
			\EndFor
		\EndFor
	\EndFor
\EndFor

\end{algorithmic}
\end{algorithm}

\begin{algorithm}[H]
\caption{Look up tables for four-card actions}\label{fourcardmoves}
\begin{algorithmic}[1]
\State \textbf{Initialize:} \textit{actionIndices}$4$ as a $13 \times 13 \times 13 \times 13$ array of zeros
\State \textbf{Initialize:} \textit{inverseIndices}$4$ as an $330 \times 4$ array of zeros
\State \textbf{Initialize:} $i = 0$
\For{$c_1=0$ to $9$}
	\State $n_1 = \min (c_1 + 3, 10)$
	\For{$c_2=c_1+1$ to $n_1$}
		\For{$c_3 = c_2+1$ to $11$}
			\State $n_2 = \min (c_3+3, 12)$
			\For{$c_4=c_3+1$ to $n_2$}
				\State \textit{actionIndices}$4[c_1, c_2, c_3, c_4] = i$
				\State \textit{inverseIndices}$4[i,:] = [c_1, c_2, c_3, c_4]$
				\State $i \mathrel{+}= 1$
			\EndFor
		\EndFor
	\EndFor
\EndFor

\end{algorithmic}
\end{algorithm}

\begin{algorithm}[H]
\caption{Look up tables for three-card actions}\label{threecardmoves}
\begin{algorithmic}[1]
\State \textbf{Initialize:} \textit{actionIndices}$3$ as a $13 \times 13 \times 13$ array of zeros
\State \textbf{Initialize:} \textit{inverseIndices}$3$ as an $31 \times 3$ array of zeros
\State \textbf{Initialize:} $i = 0$
\For{$c_1=0$ to $10$}
	\State $n_1 = \min (c_1 + 2, 11)$
	\For{$c_2=c_1+1$ to $n_1$}
		\State $n_2 = \min (c_1+3,12)$
		\For{$c_3=c_2+1$ to $n_2$}
			\State \textit{actionIndices}$3[c_1, c_2, c_3] = i$
			\State \textit{inverseIndices}$3[i,:] = [c_1, c_2, c_3]$
			\State $i \mathrel{+}= 1$
		\EndFor
	\EndFor
\EndFor
\end{algorithmic}
\end{algorithm}

\begin{algorithm}[H]
\caption{Look up tables for two-card actions}\label{twocardmoves}
\begin{algorithmic}[1]
\State \textbf{Initialize:} \textit{actionIndices}$2$ as a $13 \times 13$ array of zeros
\State \textbf{Initialize:} \textit{inverseIndices}$2$ as an $33 \times 3$ array of zeros
\State \textbf{Initialize:} $i = 0$
\For{$c_1=0$ to $11$}
	\State $n_1 = \min (c_1 + 3, 12)$
	\For{$c_2=c_1+1$ to $n_1$}
		\State \textit{actionIndices}$2[c_1, c_2] = i$
		\State \textit{inverseIndices}$2[i,:] = [c_1, c_2]$
		\State $i \mathrel{+}= 1$
	\EndFor
\EndFor

\end{algorithmic}
\end{algorithm}

\newpage

\section*{Appendix C. Details About the Training Algorithm/ Neural Network Architecture}
\label{app:trainingalg}
If the weights and biases of the neural network are contained in a vector $\theta$ then to implement the PPO algorithm we start by defining the "conservative policy iteration" loss estimator \citep{kl02}
\begin{equation}
L^{CPI}(\theta) = \hat{\mathbb{E}}_t \left[ \frac{\pi_{\theta}(a_t | s_t)}{\pi_{\theta_{old}}(a_t | s_t)} \hat{A}_t \right]
\end{equation}
where here the expectation is taken with respect to a finite batch of samples generated using the current policy parameters $\theta_{old}$. Trust region policy optimization methods maximize this loss subject to a constraint on the KL divergence between $\pi_{\theta}$ and $\pi_{\theta_{old}}$ to prevent policy updates occurring which are too large. PPO is able to achieve essentially the same thing by introducing a new hyperparameter $\epsilon \ll 1$ and instead using a clipped loss function that removes the incentive to make large policy updates. If we define $r_t(\theta) = \frac{\pi_{\theta}(a_t | s_t)}{\pi_{\theta_{old}}(a_t | s_t)}$ then PPO considers instead maximizing the following "surrogate loss function":
\begin{equation}
L^{CLIP}(\theta) = \hat{\mathbb{E}}_t \left[ \min \left(  r_t(\theta) \hat{A}_t, \text{clip} \left(  r_t(\theta), 1-\epsilon, 1+\epsilon \right)   \right) \right]
\end{equation}
We then also include a value function error term as well an entropy bonus to encourage exploration such that the final loss function to be optimized is
\begin{equation}
L(\theta) = \hat{\mathbb{E}}_t \left[ L^{CLIP}(\theta) - a_1 L^{VF}(\theta) + a_2 S[\pi_{\theta}](s_t) \right]
\end{equation}
where $a_1$ and $a_2$ are hyperparameters, $S$ is the entropy and $L^{VF} = (V_{\theta}(s_t) - V_t^{target})^2$ is the squared-error value loss. We estimate the returns and the advantages using "generalized advantage estimation" which uses the following estimate:
\begin{equation}
\hat{A}_t = \delta_t + (\gamma \lambda) \delta_{t+1} + \dots + (\gamma \lambda)^{T-t+1} \delta_{T-1}
\end{equation}
where $T$ is the number of time steps we are simulating to generate each batch of training data, $\gamma$ is the discount factor, $\lambda$ is another hyperparameter and $\delta_t = r_t + \gamma V(s_{t+1}) - V(s_t)$ (with $r_t$ being the actual reward received at time step t). 

When a batch is generated by running $N$ separate games each for $T$ time steps and the advantage estimates are made training then occurs for $K$ epochs using a minibatch size of $M$. The hyperparameters we used for our training were the following:
$N = 48, T = 20, \gamma = 0.995, \lambda = 0.95, M = 240, K = 4, a_1 = 0.5, a_2 = 0.02$ with a learning rate $\alpha=0.00025$ and $\epsilon = 0.2$ which were both linearly annealed to zero throughout the training.

\begin{figure}
\centering
\includegraphics[scale=0.7]{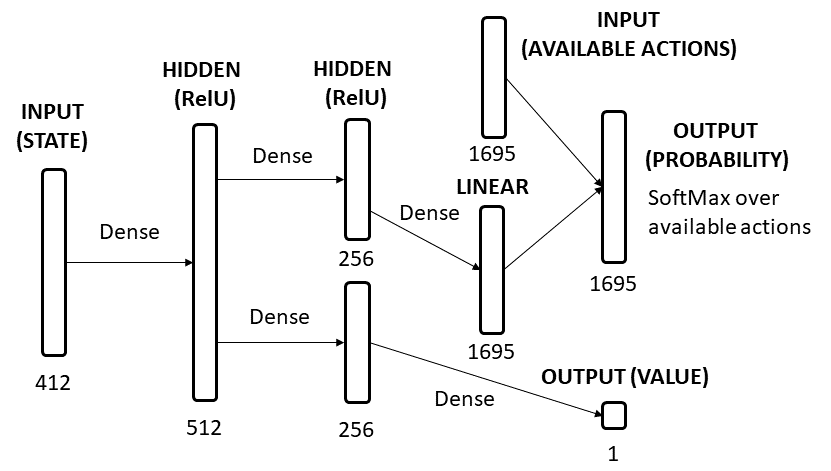}

\caption{Architecture of the neural network used. }
\label{fig:networkdiagram}
\end{figure}

In terms of the neural network architecture we used this is shown in figure \ref{fig:networkdiagram}. We have an initial shared hidden layer of 512 RelU activated units which is connected to two separate second hidden layers each of 256 RelU activated units. One of these produces an output corresponding to the estimated value of the input state whilst the other is connected to a linear output layer of 1695 units which represents a probability weighting of each potentially allowable move. This is then combined with the actually allowable moves to produce an actual probability distribution. The rationale for having a shared hidden layer is that there are likely to be features of the input state that are relevant for both evaluating the state's value as well as the move probabilities although we did not run any tests to quantify whether this is significant. All layers in the network are fully connected.

\section*{Appendix D. Results Against Human Players}
Results against seven different human players are shown in the table below. 
\begin{table}[h!]
\scriptsize
	\makebox[\textwidth][c]{\begin{tabular}{ | c | c | c | c | c | c | c | c | c |}
	\hline
	 & Player 1 & Player 2 & Player 3 & Player 4 & Player 5 & Player 6 & Player 7 & Total \\
	 \hline
	 
	 Games Played & 250 & 127 & 100 & 55 & 50 & 50 & 31 & 664 \\
	 Games Won & 68 (27.2\%) & 25 (19.7\%) & 19 (19.0\%) & 21 (38.2\%) & 5 (10.0\%) & 4 (8.0\%) & 7 (22.5\%) & 149 (22.5\%)\\
	 Final Score & -128 & -118 & -154 & 104 & -100 & -231  & -10 &  -637\\
	 Average Score & -0.51 & -0.93 & -1.54 & 1.89 & -2.00 & -4.62 & -0.32 & -0.96\\
	 Standard Error & 0.68 & 0.86 & 0.88 & 1.54 & 1.09 & 0.93 & 1.52 & 0.38\\
	 AI Scores & 51, 58, 19 & 15, -78, 181 & 73, -143, 224 & -71, -15, -18 & 86, 8, 6 & 137, 116, -22 & 124, -77, -37 & 415, -131, 353\\
	 
	AI (1) Average & 0.20 $\pm$ 0.67 & 0.12 $\pm$ 0.95 & 0.73 $\pm$ 1.04 & -1.29 $\pm$ 1.23 & 1.72 $\pm$ 1.48 & 2.74 $\pm$ 1.70 & 4.00 $\pm$ 2.07 & 0.63 $\pm$ 0.41 \\
	
	AI (2) Average & 0.23 $\pm$ 0.67 & -0.61 $\pm$ 0.89 & -1.43 $\pm$ 0.84 & -0.27 $\pm$ 1.35 & 0.16 $\pm$ 1.39 & 2.32 $\pm$ 1.75 & -2.48 $\pm$ 1.31 & -0.20 $\pm$ 0.39 \\
	
	AI (3) Average & 0.08 $\pm$ 0.65 & 1.43 $\pm$ 1.09 & 2.24 $\pm$ 1.12 & -0.33 $\pm$ 1.20 & 0.12 $\pm$ 1.29 & -0.44 $\pm$ 1.48 & -1.19 $\pm$ 1.37 & 0.53 $\pm$ 0.41 \\

	 \hline

	\end{tabular}}	
	\caption{Data from games of seven different human players vs. 3 of the trained neural networks. Standard errors on the average scores are calculated as $\sigma_m = \sigma / \sqrt{N}$ where $\sigma$ is the standard deviation of the game scores and $N$ is the number of games played.}
\end{table}

Although we only have a relatively small data set and Big 2 is a game of large variance in the scores it is clear that on the whole the neural network quite significantly outperforms the human players. Of the seven players who played only one of them finished with a positive score and this was from a relatively small number of games (Big 2 is a zero-sum game and so any negative score can be considered as a loss). If we look at the total scores of all of the human players combined we find an average score of $-0.96 \pm 0.38$ which shows that on the whole the trained neural network seems to have a significant advantage. 

\begin{figure}
\includegraphics[scale=0.5]{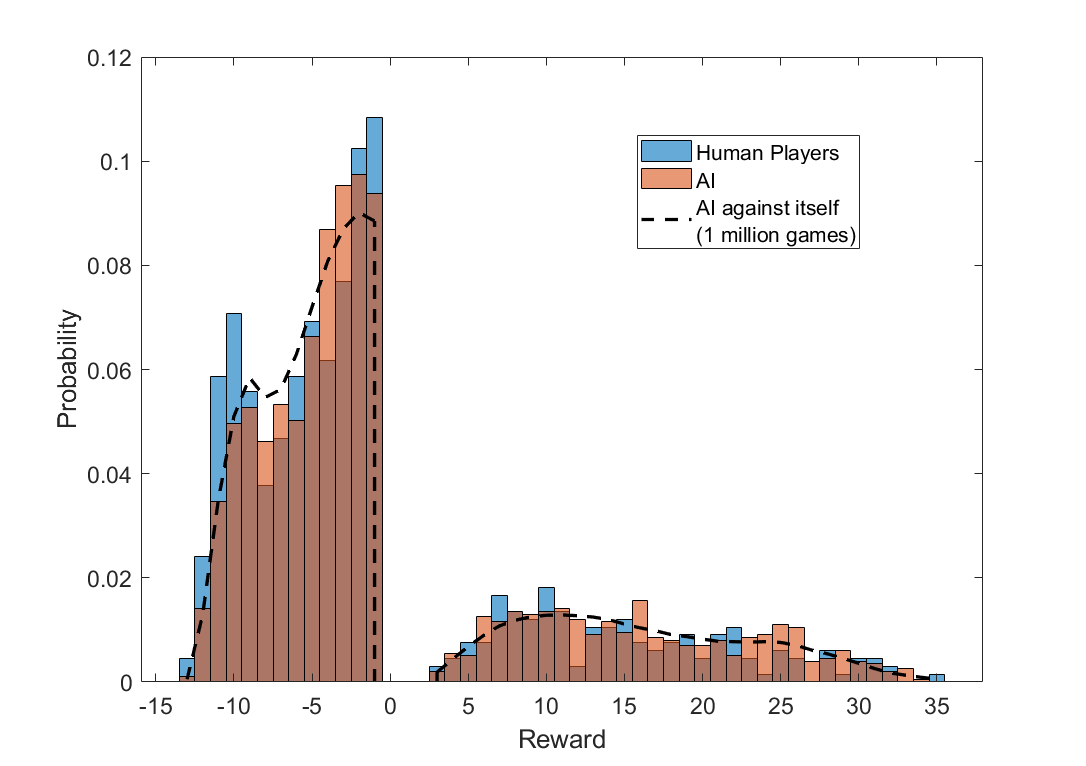}

\caption{Probability distribution of the rewards received from the games between the AI and various human players (see table 1 for a summary of results). For comparison the black line is the probability distribution from four of the fully-trained neural networks playing against themselves over 1 million games. }
\label{fig:rewardhistograms}
\end{figure}

We can also look at the probability distribution of the rewards (figure \ref{fig:rewardhistograms}) to potentially get more insight into how the neural network plays compared with the human players. One of the main differences we see is that the human players seem to find themselves left with a large number of cards more frequently than the AI does, perhaps as the AI is better able to identify situations where the chances of winning is very low and so knows just to get rid of as many cards as possible. It also seems like the AI is slightly better at ending the game early and so achieving the higher scores (which could also be the reason why human players have more cards left more often), although really we need to gather more data to be able to say anything concrete here.

\vskip 0.2in

\end{document}